\documentclass[sigconf]{acmart}

\AtBeginDocument{%
  }

\setcopyright{acmlicensed}
\copyrightyear{2026}
\acmYear{2026}
\setcopyright{cc}
\setcctype{by}
\acmConference[IUI MIRAGE 2026 workshop]{IUI MIRAGE 2026 workshop held at 31st International Conference on Intelligent User Interfaces}{March 23, 2026}{Paphos, Cyprus}
\acmBooktitle{IUI MIRAGE 2026 workshop held at 31st International Conference on Intelligent User Interfaces}
\acmDOI{}
\acmISBN{}

\acmConference[IUI '26]{Joint Proceedings of the ACM IUI Workshops 2026, Misleading Impacts Resulting from AI Generated Explanations}{March 23--26, 2026}{New York City, NY, USA}

\usepackage{tcolorbox}
\usepackage{siunitx}
\usepackage{wrapfig}
\makeatletter
\newlength{\originalauthorbxwd}
\AtBeginDocument{%
  \setlength{\originalauthorbxwd}{\author@bx@wd}
  \setlength{\author@bx@wd}{0.85\originalauthorbxwd}  
}

\renewcommand\@authorfont{\fontsize{10pt}{13pt}\selectfont\sffamily}
\renewcommand\@affiliationfont{\normalsize\normalfont}
\makeatother

\begin{document}

\title{Not Too Short, Not Too Long: How LLM Response Length Shapes People’s Critical Thinking in Error Detection}

%
%



\author{Natalie Friedman}
\email{natalie.friedman@sap.com}
\affiliation{
  \institution{BTP Innovation, SAP}
  \city{Palo Alto}
  \state{CA}
  \country{USA}
}

\author{Adelaide Nyanyo}
\email{Adelaide.nyanyo@sap.com}
\affiliation{
  \institution{BTP Innovation, SAP}
  \city{Palo Alto}
  \state{CA}
  \country{USA}
}

\author{Kevin Weatherwax}
\email{Kevin.weatherwax@sap.com}
\affiliation{
  \institution{BTP Innovation, SAP}
  \city{Palo Alto}
  \state{CA}
  \country{USA}
}

\author{Lifei Wang}
\email{lifei.wang@sap.com}
\affiliation{
  \institution{BTP Innovation, SAP}
  \city{Palo Alto}
  \state{CA}
  \country{USA}
}

\author{Chengchao Zhu}
\email{chengchao.zhu@sap.com}
\affiliation{
  \institution{BTP Innovation, SAP}
  \city{Palo Alto}
  \state{CA}
  \country{USA}
}

\author{Zeshu Zhu}
\email{zeshu.zhu@sap.com}
\affiliation{
  \institution{BTP Innovation, SAP}
  \city{Palo Alto}
  \state{CA}
  \country{USA}
}

\author{S. Joy Mountford}
\email{joy.mountford@sap.com}
\affiliation{
  \institution{BTP Innovation, SAP}
  \city{Palo Alto}
  \state{CA}
  \country{USA}
}

\renewcommand{\shortauthors}{Friedman et al.}

\begin{abstract}

Large language models (LLMs) have become common decision-support tools across educational and professional contexts, raising questions about how their outputs shape human critical thinking.  
Prior work suggests that the amount of AI assistance can influence cognitive engagement, yet little is known about how specific properties of LLM outputs (e.g., response length) impacts users’ critical evaluation of information.  
In this study, we examine whether the length of LLM responses shapes users’ accuracy in evaluating LLM-generated reasoning on critical thinking tasks, particularly in interaction with the correctness of the LLM’s reasoning.  
To begin evaluating this, we conducted a within-subjects experiment with 24 participants who completed 15 modified Watson--Glaser critical thinking items, each accompanied by an LLM-generated explanation that varied in length and correctness.  
Mixed-effects logistic regression revealed a strong and statistically reliable effect of LLM output correctness on participant accuracy, with participants more likely to answer correctly when the LLM’s explanation was correct.  
Response length appeared to moderated this effect: when the LLM output was incorrect, medium-length explanations were associated with higher participant accuracy than either shorter or longer explanations, whereas accuracy remained high across lengths when the LLM output was correct.  
Together, these findings suggest that response length alone may be insufficient to support critical thinking, and that how reasoning is presented—including a potential advantage of mid-length explanations under some conditions—points to design opportunities for LLM-based decision-support systems that emphasize transparent reasoning and calibrated expressions of certainty.

\end{abstract}




\maketitle


\section{Introduction}

Usage of large language model (LLM) tools has fast become ubiquitous in assisting with learning, decision making, and a wide variety of tasks in personal, scholastic, and professional settings.
This rapid expansion of LLM use has outpaced our understanding of how these tools affect human cognition and judgment, even as they are increasingly embedded within everyday workflows.
While LLMs are effective at quickly generating solutions and suggestions, people still must evaluate the quality of those suggestions and estimate appropriate trust to decide how, or whether, to act on them.
As a result, LLMs do not replace human reasoning so much as reshape when, where, and how cognitive effort is applied.
However, we do not yet fully understand how exposure to large volumes of confidently presented text, particularly when that text contains errors, influences human judgment and decision making.
LLM outputs may appear coherent and authoritative at first glance, even when they are incomplete, misleading, or incorrect.
This raises broader questions about how LLM-assisted interactions shape critical thinking.
More specifically, does the presence of LLM assistance affect how people encode information, allocate cognitive effort, and evaluate decisions?

Prior research on LLMs has largely focused on properties of the models themselves, including hallucinations, bias, error, misinformation, and trustworthiness~\cite{badyal2023intentional}, ~\cite{yadkori2024believe}, ~\cite{kamoi2024evaluating}, ~\cite{chen2023can}.
Although this work is essential, it often treats users as passive recipients of LLM output rather than active evaluators of information.
Comparatively little research has examined how characteristics of LLM responses themselves, such as information structure (e.g., paragraphs versus bullet points), reasoning depth, or response length, shape information assimilation and critical engagement.

Many contemporary LLMs present an answer followed by an explanation and then restate the answer, with explanations commonly framed as mechanisms for improving transparency, trust, and understanding.  
Although explanations may support comprehension, they can also introduce risks, including inflated trust driven by overconfident reasoning, distorted mental models, and reduced critical scrutiny.  
Such risks may be especially pronounced when explanations are lengthy, fluent, or incorrect.  

In LLM interfaces, answers are often accompanied by explanations. These explanations are intended to improve transparency and support appropriate trust. However, explanation properties such as length may also influence user reasoning in unintended ways.

In this work, we examine LLM output features that may exacerbate explainability pitfalls, focusing on LLM output length and correctness.  
We foreground how these LLM output characteristics influence human reasoning.  
By doing so, this work aims to contribute to research on human--LLM interaction and has implications for how LLMs are used in practice, as well as for the design of LLM output architectures and user interfaces.

\section{Background}

Understanding how people think has become increasingly important as AI systems play a larger role in everyday reasoning. In this section, we outline foundational theories and measurements of critical thinking and describe how AI assistance can both support and hinder critical thought. Finally, we survey current AI-driven decision-support tools in enterprise settings to illustrate how large language models (LLMs) are redesigning the presentation of information and, in turn, influencing human judgment.

\subsection{Critical Thinking: Conception \& Measurement}
Critical thinking is a core facet of human performance in nearly all scholastic and vocational settings~\cite{bernard2008exploring,lai2011critical,halpern2013thought}.
Critical thinking is often misunderstood and used as a catch-all to describe intelligence.
This is because it is an umbrella term---or summation---of peoples ability to practice deductive reasoning, analyze arguments, understand likelihoods, make decisions, problem solve, and engage in creativity ~\cite{halpern2013thought}.
For this reason, people with strong critical thinking skills are highly sought after in most professional settings~\cite{ahuna2014new,halpern2013thought}.
Yet critical thinking itself is difficult to teach~\cite{willingham2007critical}.

For a working definition, critical thinking is an outcome of a variety of meta-thinking or meta-cognitive skills and strategies~\cite{willingham2007critical,halpern2013thought}. 
However, it requires more than just ``thinking about thinking''~\cite{halpern2013thought}.
In effect, critical thinking requires people to consider, carefully, not just a decision or conclusion that is being made, but the steps by which people arrive at it and the logic, reason, or facts that support it~\cite{halpern2013thought}.
More specifically, is about moving beyond retention of rote information and learning how to masterfully apply said knowledge~\cite{halpern2013thought}.
It is understanding not just \textit{how} \textit{you} arrived at a thought but \textit{how} \textit{others} may arrive at your same perspective, as well as alternate or even opposing perspectives. 
A common example in espousing the value of critical thinking stems from legal professions (e.g., trial lawyers) where the success often derives from someones ability to carefully map out, and then lead others, to very specific points of view while also predicting likely deviations of thought which could lead people (e.g., a jury or judge) to undesirable outcomes of position (e.g., finding someone guilty who is, in fact innocent).
Similarly, in business settings it can be applied towards predicting likely customer or client behavior or considering the next moves of market competitors.

Within educational and professional research, critical thinking has been examined closely, at length, and for many decades~\cite{halpern2013thought}.
The first, and still foremost, measurement and conception of critical thinking---as well as its underlying facets---comes from work by \citet{Watson1980}.
The Watson-Glaser Critical Thinking Assesment (WGCTA) is still widely used for applicants to jobs in various fields and to assess outcomes in educational interventions.
The WGCTA breaks down critical thinking into five primary facets; making inferences, recognizing assumptions, drawing deductive conclusions, interpreting information, and evaluating arguments~\cite{Watson1980}.
In the present work we used a modified 15 question version of the WGCTA combined with an LLM output to each in order to assess how LLM assistance impacted participants' critical thinking (See Sec. \ref{sec:Methods} for more details).

\subsection{AI assistance impact on critical thinking}

Critical thinking relies on effective working memory through a person's ability to encode and store information. 
The encoding-storage paradigm is defined in educational psychology, as a two-step process in learning: encoding ``the initial intake and processing of information,'' and storage, ``the retention of that information over time''~\cite{kiewra1989review}.
Backed up by the encoding-storage paradigm, researchers in human-computer interaction have hypothesized and found that more encoding with storage will lead to better information retention. 
But what if encoding is hindered?
\citet{chen2025more} found that too much AI assistance can reduce encoding. 
They studied the impact of AI assistance on cognitive engagement through the lens of the \textit{AI Assistance Dilemma} which describes the balance between assistance and autonomy in learning without creating a dependency \cite{koedinger2007exploring}. 
In particular, \citet{chen2025more} reported that while people preferred higher amounts of assistance because of the lower cognitive effort, intermediate assistance actually yielded better test results.

Assistance that reduces the need to generate reasoning may lower effort, but longer explanations may also increase processing demands due to greater reading time and working memory load. Thus, explanation length may exert competing influences on critical thinking performance with assistance.

Education researchers \citet{liu2025generative} reviewed 15 papers on using LLMs to learn languages and it's impact on critical thinking. More specifically, they were working to understand if LLMs were hindering or helping people's ability to critically think in learning English as a Foreign Language. They found ``66.67\% of studies reported generative AI tools and LLMs’ positive role in CT, while 33.33\% of studies reported its negative role in CT.'' \cite{liu2025generative}.

\subsection{AI tools for Decision Support: A Review of Current Technology}
In enterprise and applied settings, decision making is a central part of daily work. Good decisions depend on critical thinking, tools that support decision making thus play an important role in practice.

There are an increasing number of AI tools that aim to support decision-making in professional workflows. 
For example, Microsoft 365 Copilot provides a series of functions across its products: in Excel it identifies patterns, generates formulas, and surfaces anomalies; in Teams it provides meeting analysis and summarizes action items. 
Google Workspace takes a similar approach. 
Gemini for Workspace provides decision-memo generation, summaries of lengthy documents, and auto-extracts tasks and risks from emails. 
In both ecosystems, the AI system supports decision-making by highlighting what's important, giving an initial interpretation of the information, and providing quick summaries that often become the starting point for how user make their decisions.


AI is also becoming part of how teams make decision together. 
Slack AI works on turning long messages threads and busy channels into short summaries that are easier to catch up on. 
It can point out open questions, highlight tasks that people mentioned and tag messages that need attention. 
This helps teams stay aligned without needing to read through everything. 
Tools like Notion AI offer similar support by turning scattered notes and updates into a clearer picture of a project to better facilitate group decision making.

Overall, these tools show how AI is now involved in many everyday decision making process at work. 
Across productivity tools and team platforms, AI often provides the first interpretation of information that people rely on. 
This makes LLM output an important factor in how decisions are formed and how users approach critical thinking.

In parallel with these enterprise tools, many people also use general-purpose LLMs, such as ChatGPT, Gemini, and Claude, directly in their daily work. These models are often used to think through problems, compare options and get a quick first take, even outside of formal enterprise systems. As a result, they serve as another form of decision support that is widely used in practice and relevant to study more closely.

\subsection{Hypotheses}\label{Sub:H1H2}

Based on prior work on critical thinking, we hypothesize that LLM output length will influence users’ critical thinking performance.  
Longer LLM outputs may support deeper encoding and processing than direct answers, potentially increasing critical engagement.  
For example, \citet{chen2025more} found that moderate LLM involvement in note-taking improved engagement and test accuracy, whereas excessive involvement reduced performance.  
Accordingly, we hypothesize that LLM output length may impact accuracy on a critical thinking task.  
We note that prior work examined user engagement rather than critical thinking, and these constructs may not yield identical outcomes.

\textbf{Hypothesis 1:} Participants will demonstrate higher accuracy when evaluating medium-length LLM outputs on critical thinking tasks than when evaluating shorter or longer outputs.

\textbf{Hypothesis 2:} The correctness of the LLM outputs may influence participants’ critical thinking accuracy, such that correct LLM outputs will be associated with higher participant accuracy than incorrect outputs.








\section{Methods}\label{sec:Methods}
To test our hypotheses (see Sec.~\ref{Sub:H1H2}), we conducted a within-subjects experiment in which 24 participants (46\% male, 54\% female; ages 24--65, \textit{M} = 42, \textit{SD} = 12) evaluated LLM-generated outputs to critical thinking questions that varied in correctness and output length~\cite{Watson1980}.  
Participants were compensated \$70 per hour and were primarily based in the US and UK. Participants were primarily based in the US and UK and held professional roles across finance, consulting, healthcare, public administration, technology, logistics, and higher education. Roles ranged from entry-level positions to managerial, senior, and partner-level positions. Most participants regularly engaged in analytical or decision-making responsibilities within their roles.

\begin{figure}
    \centering
   \begin{tcolorbox}
\textbf{Watson Glaser Statement:}
Freedom of speech is an important right in modern society. It means having the right to speak one's mind regardless of what one's opinion might be. Any country that tries to dictate public opinions by law cannot be called a democracy.\\

\textbf{Watson Glaser Proposed Assumption:}
Freedom of speech hasn't always been an important part of a democracy. The task is to evaluate whether the proposed assumption is necessary for the statement to be true.\\

\textbf{LLM Statement:}
Freedom of speech is an important right in modern society. It means having the right to speak one's mind regardless of what one's opinion might be. Any country that tries to dictate public opinions by law cannot be called a democracy.\\

\textbf{LLM Proposed Assumption:}
Freedom of speech hasn't always been an important part of a democracy.\\

\textbf{LLM Analysis:}
An assumption is something that must be true for the statement to hold true. Let's test this:
The statement asserts that freedom of speech is important in modern society and essential for democracy today (i.e., if a country restricts speech, it's not a democracy).

The proposed assumption talks about the historical relationship between freedom of speech and democracy---specifically suggesting that freedom of speech was not always important in democracies of the past.
But the statement says nothing about history. It talks about the current importance of freedom of speech in modern democracies. So:
Even if the assumption were false (i.e., freedom of speech was always important in democracies), the statement could still be true.
Therefore, the proposed assumption is not necessary for the statement to be true.\\

\textbf{LLM Conclusion:}
No, the proposed assumption is not necessary for the statement to be true.
\end{tcolorbox}
\vspace{4pt}
    \caption{Example LLM output for a Recognizing Assumptions item from the Watson--Glaser scale.}
    \vspace{-8pt}
    \label{LLM output}
\end{figure}



\subsection{Task}
Participants completed 15 questions from a modified Watson--Glaser scale \cite{Watson1980}, which assesses critical thinking across five categories (inference, recognition of assumptions, deduction, interpretation, and evaluation of arguments).  

Each LLM output consisted of (1) a step-by-step analysis and (2) a final yes/no conclusion. Participants viewed both components together and were asked to evaluate the correctness of the LLM’s overall outputs, including its reasoning and final answer. Only the LLM output varied in word count. The original Watson–Glaser prompts remained identical across conditions.

We selected the Watson–Glaser Critical Thinking Appraisal because it provides a validated, widely used measure of analytical reasoning and error detection across inference, assumption recognition, deduction, and argument evaluation. These skills closely align with the task required participants to critically evaluate the correctness of LLM-generated reasoning.

Participants were not asked to independently solve the Watson–Glaser items; instead, they evaluated the correctness of an LLM-generated output to each item. For each question, participants read an LLM-generated output and judged its correctness (see Figure~\ref{LLM output} for an example).  
LLM outputs were generated in August 2025 using ChatGPT’s default model (GPT-5).  
Participants saw three questions from each category in randomized order.  
Of the 15 LLM outputs, eight were correct and seven contained planted errors.  For incorrect conditions, errors were introduced in the final conclusion while the preceding reasoning steps were left intact.
LLM output length varied from 42 to 150 words (\textit{M} = 82.2, \textit{Mdn} = 70).

\subsection{Measures}
We measured participants’ judgments of LLM output correctness and collected written rationales for those judgments.  
Correctness was assessed by asking participants, “Is the LLM’s response correct?”.  
LLM outputs were coded as correct when participants agreed with a correct LLM output or disagreed with an incorrect LLM output, and as incorrect otherwise.  
Following prior work on cognitive engagement and correctness~\cite{chen2023can}, this measure captures whether participants align with or challenge the LLM’s output when forming their judgments.  
Participants also provided brief written explanations of their judgments, which were collected for future qualitative analysis.

To examine the effect of LLM output length, outputs were categorized into short, medium, and long conditions based on word count. Rather than using arbitrary thresholds, we used a data-driven, quantile-based approach and divided the distribution of LLM output lengths into tertiles (lowest third = short, middle third = medium, highest third = long). Across items, LLM outputs ranged from 42–150 words (M = 82.2, Mdn = 70).

\section{Results}\label{Sec:Results}

Our hypotheses (see Sec.~\ref{Sub:H1H2}) stated that (1) participants would exhibit higher critical thinking accuracy when evaluating medium-length LLM responses than when evaluating shorter or longer responses, and (2) the correctness of the LLM response would influence participants’ critical thinking accuracy, such that correct LLM outputs would be associated with higher participant accuracy than incorrect outputs.  
To evaluate these hypotheses, LLM response lengths were categorized into short, medium, and long groups using a data-driven, quantile-based approach based on the distribution of word counts, rather than arbitrary thresholds.  
We found support for both hypotheses.
Participant accuracy refers to correctly evaluating the LLM’s response (agreeing when correct and rejecting when incorrect).


\begin{table*}[t]
\centering
\footnotesize
\begin{tabular}{lccc}
\toprule
 & Short & Medium & Long \\
\midrule
LLM incorrect & 0.245 [0.127, 0.422] & 0.542 [0.362, 0.713] & 0.307 [0.169, 0.491] \\
LLM correct   & 0.715 [0.533, 0.847] & 0.811 [0.722, 0.876] & 0.795 [0.542, 0.927] \\
\bottomrule
\end{tabular}
\caption{Model-estimated probability of a correct participant response by LLM output correctness and output length (95\% CI).}
\label{tab:predicted_prob_correct}
\end{table*}

\subsection{Main Effects}

To examine whether the correctness of an LLM response influenced participant performance, we analyzed response accuracy using a mixed-effects logistic regression.  
Because response accuracy was binary and responses were repeated across questions, we analyzed the data using mixed-effects logistic regression, consistent with methodological recommendations for categorical outcome data~\cite{jaeger2008categorical}. 
In addition to fixed effects of LLM output correctness, response length, and their interaction, the model included a random intercept for question.  
This allowed each question to have its own baseline probability of being answered correctly.  

Statistical evidence for effects of interest was evaluated at the model level via estimated regression coefficients, such that the influence of LLM correctness, output length, and their interaction was assessed simultaneously within a single model.  
To aid interpretation of these effects, model-estimated predicted probabilities and corresponding 95\% confidence intervals were computed for each condition.
Results showed a strong main effect of LLM output correctness.  
More specifically, when the LLM output was correct, predicted participant accuracy ranged from approximately 71\% to 81\% across LLM output lengths.  
When the LLM output was incorrect, predicted accuracy was substantially lower, ranging from approximately 25\% to 54\% depending on output length.  

Across all length conditions, participants were markedly more likely to answer correctly when the LLM output itself was correct.
These findings indicate that responses were impacted by the accuracy of the LLM output, highlighting the potential for both performance support when the model is correct and error propagation when the model is incorrect.

The mixed-effects model also revealed a main effect of LLM output length on participant accuracy.  
Collapsing across LLM correctness, medium-length LLM outputs were associated with higher overall accuracy than short outputs, while long outputs did not yield a comparable benefit. This pattern indicates that output length influenced participant performance, but not in a simple monotonic manner.  

When the LLM output was incorrect, the model estimated a 54\% probability of a correct participant response for medium-length outputs, compared to 25\% and 31\% for short and long outputs, respectively.  
When the LLM output was correct, medium-length outputs were also associated with the highest accuracy (81\%), with lower accuracy observed for short and long outputs.  
See Table~\ref{tab:predicted_prob_correct}.  
Overall, accuracy did not increase monotonically with LLM output length; instead, medium-length outputs were associated with the highest predicted accuracy.

\subsection{Interaction Effect}

Importantly, the analysis revealed a clear interaction between LLM output correctness and output length.  
When the LLM output was incorrect, participant accuracy varied substantially by LLM output length, with medium-length outputs yielding markedly higher accuracy than either short or long outputs.  
In contrast, when the LLM output was correct, participant accuracy was relatively high across all output lengths, with only modest differences between short, medium, and long outputs.
In plain terms, when incorrect LLM outputs were relatively short or long, participants were more likely to select the same incorrect answer as the LLM.  
When incorrect LLM outputs were of medium length, participants were more likely to answer correctly despite the LLM being wrong.  
This pattern reflects differences in observed response accuracy and does not directly measure participants’ underlying cognitive processes.

\begin{wrapfigure}{r}{0.48\linewidth}
    \vspace{-0.5\baselineskip}
    \centering
    \includegraphics[width=\linewidth]{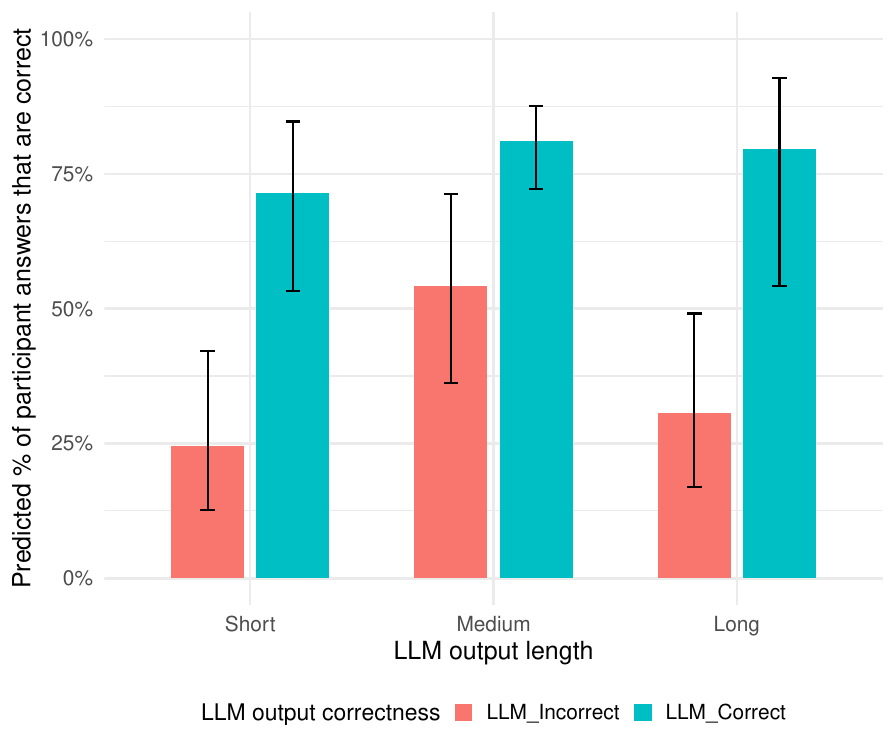}
   \caption{Predicted participant accuracy by LLM output correctness and output length (95\% CI).}
    \label{fig:correct_x_length}
    \vspace{-0.6\baselineskip}
\end{wrapfigure}

\section{Discussion}


Our findings indicate that LLM output length should be treated as an intentional design choice rather than a byproduct of LLM output. 

Contrary to the assumption that more detailed outputs would support better reasoning, longer outputs did not actually improve critical thinking performance.  

Medium-length LLM outputs, by contrast, supported a potential “sweet spot”, where users receive sufficient structure to engage with the reasoning without being overwhelmed or overly influenced. These outputs were associated with better error detection when the LLM was incorrect, while maintaining high accuracy when the LLM was correct. 

These results demonstrate the risk of defaulting to longer outputs and suggest \textbf{that output length should be intentionally constrained unless additional detail is requested.}

Finally, an early examination of participants' open-ended responses showed that explanation structure matters in addition to length. Participants often trusted the LLM’s step-by-step reasoning even when its final conclusion was incorrect, and tended to defer to the conclusion rather than scrutinize the reasoning. This suggests that tightly coupling reasoning and conclusions may lead to over-trust. Structurally separating the many steps of logic and final conclusions may better support critical evaluation by making inconsistencies more visible.




\subsection{Limitations}

The original Watson--Glaser scale consists of 40 questions completed within 30 minutes.  
For this study, we reduced the set to 15 questions and included LLM-generated outputs in order to limit participant burden and session length.  
Future work could more directly examine cognitive load using established subjective or physiological measures.

Performance varied across questions, reflecting baseline differences in item difficulty that were explicitly modeled via a random intercept for questions (see Sec. \ref{Sec:Results}).  
By accounting for question-level variability in this way, our analysis avoids attributing differences in baseline accuracy to the experimental manipulations of LLM correctness or output length.

Participants evaluated whether the LLM’s output was correct rather than answering the underlying critical thinking question directly.  
This additional evaluative step may introduce different cognitive demands than traditional critical thinking assessments, and our binary correct/incorrect measure reflects participants’ judgments of the LLM rather than their standalone reasoning ability.  
While this design choice was intentional, it limits our ability to assess how participants would have performed on the same questions without LLM support.  
Future work could include a control condition without LLM-generated outputs to more directly compare critical thinking performance with and without AI assistance.

Output length was not manipulated independently of reading time. Longer outputs necessarily required more time and effort to read, meaning that time-on-task and cognitive load were intrinsically coupled with our length manipulation. As a result, we cannot determine whether observed differences in performance stemmed from the amount of information presented, the time spent processing it, or related factors such as fatigue or attentional decline. Future work should directly measure reading time and cognitive load (e.g., by time-limiting responses or logging dwell time) to better isolate the mechanism by which output length influences critical evaluation.

Open-ended responses revealed several consistent patterns in how participants engaged with the LLM’s outputs. Many participants noted cases where they perceived the LLM’s step-by-step reasoning as logically sound but judged its final yes/no conclusion as inconsistent or contradictory with that reasoning (e.g., “the breakdown seems fine but then there is a switch at the end”). However, we did not do a thorough analysis to check for order effects and hope to analyze this in the future. In the quantitative analysis, people generally followed what the LLM was suggesting. This points to the importance of triangulating methods of both subjective and observation.

\subsection{Future Work}

To address these limitations and extend our understanding of AI's impact on critical thinking, we see some research directions for future work with larger sample sizes. 

\textbf{Cross-Model Comparative Studies: }
Our study only tested on a single LLM. Future study should expand to compare outputs across more LLMs from different developers and regions. Different models may employ distinct reasoning patterns and linguistic styles that could differentially impact user critical thinking.

\textbf{Multilingual Studies: }
Our current study is limited to English-speaking participants. Future work should translate the Watson-Glaser assessment into multiple languages and recruit participants from diverse linguistic backgrounds. The expansion is critical because languages could vary in information density, communication norms differ across cultures, and different languages may elicit varying output length from LLMs. By conducting parallel studies across languages and models, we can determine whether our findings represent universal patterns or context-specific phenomena.

\textbf{Systematic Output Length Manipulation: }
When we prompted the LLM, the output lengths were not evenly distributed across outputs. Future studies should systematically control output length by creating small, medium, and large output conditions (for example, 50, 100, 150 word counts) with while maintaining consistent information content. By grouping these we can better assess a relationship between word count and critical thinking.















\section{Conclusion}

This study examined whether the length and correctness of LLM-generated outputs influences critical thinking accuracy. Our results show that output length alone does not reliably improve performance; instead, participant accuracy was driven by the correctness and internal consistency of the LLM output. Participants were significantly more likely to answer correctly when the model’s explanation was correct, while incorrect explanations often propagated errors. These findings highlight an important explainability pitfall: explanations can both support and hinder human reasoning depending on their quality, not their verbosity. Designing safer AI explanations therefore requires prioritizing reasoning clarity, consistency, and accurate expressions of certainty rather than longer outputs.


\bibliographystyle{ACM-Reference-Format}
\bibliography{bibliography}


\end{document}